\documentclass[lettersize,journal]{IEEEtran}
\usepackage{amsmath,amsfonts}
\usepackage{algorithmic}
\usepackage{algorithm}
\usepackage{array}
\usepackage[caption=false,font=normalsize,labelfont=sf,textfont=sf]{subfig}
\usepackage{textcomp}
\usepackage{stfloats}
\usepackage{url}
\usepackage{verbatim}
\usepackage{graphicx}
\usepackage{cite}
\usepackage{xspace}
\usepackage{xcolor}
\usepackage{amssymb}
\usepackage{multirow}
\usepackage{booktabs}
\usepackage[hidelinks]{hyperref}

\newcommand{\eg}{e.g.\xspace}
\newcommand{\ie}{i.e.\xspace}

\begin{document}

\title{ESOM: Efficiently Understanding Streaming Video Anomalies \\ with Open-world Dynamic Definitions}

\author{Zihao Liu, Xiaoyu Wu, Wenna Li, Jianqin Wu, Linlin Yang
\thanks{Zihao Liu, Xiaoyu Wu, Wenna Li, Jianqin Wu, Linlin Yang are with the State Key Laboratory of Media Convergence and Communication, Communication University
of China, Beijing, China (e-mail: liuzihao@cuc.edu.cn; wuxiaoyu@cuc.edu.cn; liwenna@cuc.edu.cn; wujianqin@mails.cuc.edu.cn; lyang@cuc.edu.cn).}%
\thanks{Corresponding author: Xiaoyu Wu.}
\thanks{Manuscript received April 19, 0000; revised August 16, 0000.}}

\markboth{Journal of \LaTeX\ Class Files,~Vol.~14, No.~8, August~2021}%
{Shell \MakeLowercase{\textit{et al.}}: A Sample Article Using IEEEtran.cls for IEEE Journals}

\IEEEpubid{0000--0000/00\$00.00~\copyright~2021 IEEE}

\maketitle

\begin{abstract}
Open-world video anomaly detection (OWVAD) aims to detect and explain abnormal events under different anomaly definitions, which is important for applications such as intelligent surveillance and live-streaming content moderation. 
Recent MLLM-based methods have shown promising open-world generalization, but still suffer from three major limitations: inefficiency for practical deployment, lack of streaming processing adaptation, and limited support for dynamic anomaly definitions in both modeling and evaluation. 
To address these issues, this paper proposes ESOM, an efficient streaming OWVAD model that operates in a training-free manner. ESOM includes a Definition Normalization module to structure user prompts for reducing hallucination, an Inter-frame-matched Intra-frame Token Merging module to compress redundant visual tokens, a Hybrid Streaming Memory module for efficient causal inference, and a Probabilistic Scoring module that converts interval-level textual outputs into frame-level anomaly scores. 
In addition, this paper introduces OpenDef-Bench, a new benchmark with clean surveillance videos and diverse natural anomaly definitions for evaluating performance under varying conditions. 
Extensive experiments show that ESOM achieves real-time efficiency on a single GPU and state-of-the-art performance in anomaly temporal localization, classification, and description generation.
The code and benchmark will be released at \href{https://github.com/Kamino666/ESOM_OpenDef-Bench}{github.com/Kamino666/ESOM\_OpenDef-Bench}.
\end{abstract}


\section{Introduction}

Video anomaly detection (VAD) aims to detect video events that deviate from what is expected \cite{VAD-review-wupeng}, which plays an important role in intelligent surveillance and live-streaming content moderation. 
Open-World Video Anomaly Detection (OWVAD) is a promising research direction, with existing methods exploring generalization to unseen scenarios \cite{ovvad,anomize,evidential_open_world}, explainability \cite{LAVAD,Holmes-VAU}, and adaptation to dynamic anomaly definitions \cite{liu2025lagovad}. 
However, existing methods fail to jointly integrate these advantages, and often suffer from inefficiency and a lack of streaming video inference capability, making them less practical for open-world video anomaly detection.

\begin{figure}[!t]
    \centering
    \includegraphics[width=0.98\columnwidth]{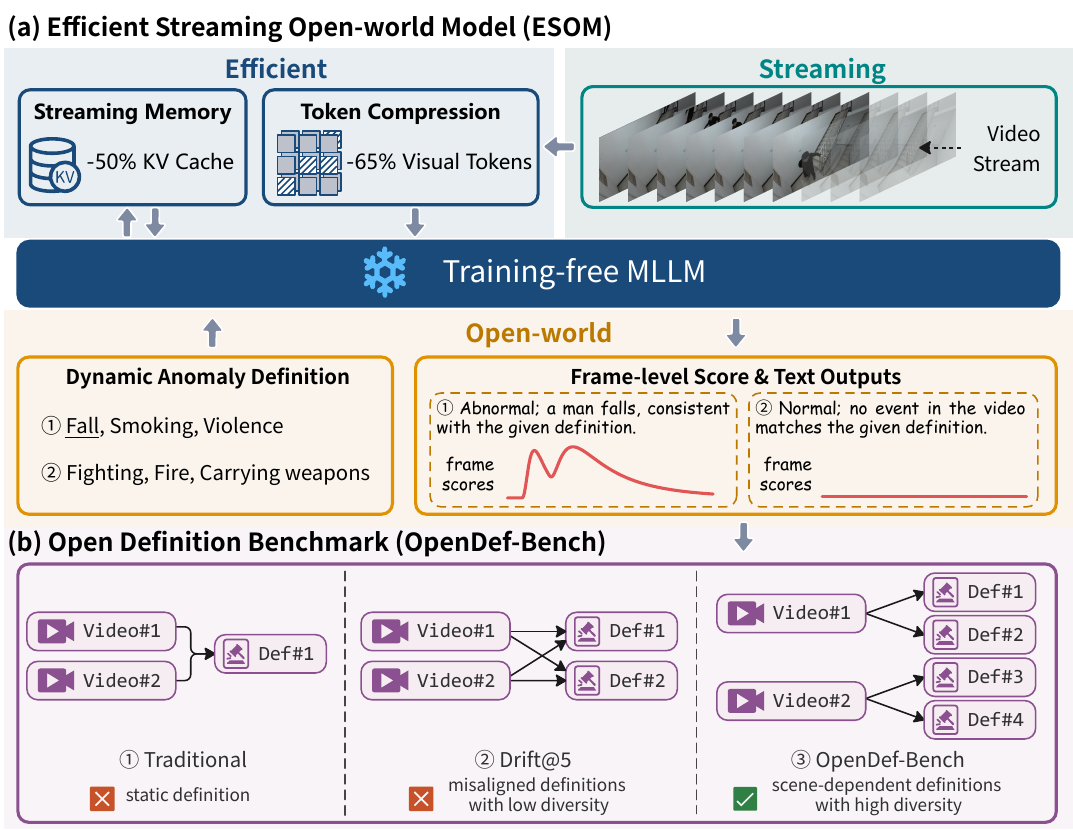}
    \caption{Motivation of ESOM and OpenDef-Bench. (a) ESOM addresses three key challenges in OWVAD: efficiency, streaming processing, and ability of generating frame-level anomaly scores and textual explanations under dynamic anomaly definitions. (b) OpenDef-Bench provides diverse, scene-dependent anomaly definitions. In contrast, Traditional settings \cite{ovvad} rely on static definitions, while recent Drift@5 \cite{liu2025lagovad} suffers from limited diversity and weak alignment with real-world scenarios.}
    \label{fig:motivation}
    \vspace{-1em}
\end{figure}

Although the concept of \textit{open-world} in OWVAD has not yet been unified, existing studies generally characterize it from generalization and explainability perspectives.
\textbf{For generalization}, the challenges lie in adaptation to unseen scenarios and dynamic anomaly definitions.
First, generalizing to unseen scenarios is difficult for methods based on small models with task-specific fine-tuning \cite{vadclip,GlanceVAD,chen2025dctformer}, as their capacity is constrained by the limited scale and diversity of existing VAD training data, while recent training-free LLM-based methods \cite{LAVAD,vera} can better exploit large-scale pre-trained knowledge and have shown stronger zero-shot generalization.
\IEEEpubidadjcol
Second, generalizing to dynamic anomaly definitions raises a fundamental question concerning what should be regarded as an anomaly.
Semi-supervised and unsupervised approaches \cite{one-class-2017,one-class-wu2020} typically treat events that deviate from the distribution of normal data as abnormal, whereas supervised and weakly supervised methods \cite{vadclip,GlanceVAD} define anomalies according to the labeled categories in the dataset.
Both paradigms implicitly assume static anomaly definitions, while in practice abnormality often varies with user requirements (\eg, periods, policies, and scenes).
A pioneering work \cite{liu2025lagovad} takes a first step by jointly modeling videos and anomaly definitions. However, due to limited training data, its performance degrades significantly on unseen scenes.
\textbf{For explainability}, traditional methods \cite{liu2025lagovad,GlanceVAD,ovvad} lack language generation ability and therefore cannot provide detailed anomaly analysis.
By contrast, some LLM-based methods \cite{HAWK,vadR1,vauR1,PANDA} can readily generate textual outputs, but struggle to produce fine-grained frame-level anomaly scores.
Together, these challenges motivate a unified framework that addresses the key goals of OWVAD, as shown in Fig.~\ref{fig:motivation}a.

To enable OWVAD, Multi-modal LLMs (MLLMs) play a crucial role. However, another key obstacle is whether the model can support efficient streaming inference. In surveillance and live-streaming scenarios, both causality and real-time inferencing are essential: the former requires the model to rely only on previous infomation, while the latter requires the processing speed to keep up with the video stream.
However, existing training-free methods \cite{LAVAD,vera,PANDA} build a complex pipeline with agents or multiple models, which fail to meet real-time requirements, and single-model methods \cite{vadR1,vauR1,HAWK} are not trained in a streaming manner.
Empirically, the main efficiency bottleneck comes from visual tokens, especially as historical information accumulates in streaming inputs. 
To address this, this paper adopts a simplified framework and proposes two plug-and-play modules: one to compress spatiotemporal redundancy in visual tokens and the other to efficiently manage and reuse previous memory, both tailored to streaming inference.

Beyond model design, existing evaluation methods are still unable to effectively measure performance under dynamic anomaly definitions. 
As shown in Fig.~\ref{fig:motivation}b, prior works \cite{ovvad,anomize} adopt evaluation with a single, fixed anomaly definition, which fails to assess performance under dynamic definitions. 
The Drift@5 protocol proposed in \cite{liu2025lagovad} instead evaluates with five different anomaly definitions. However, the diversity of these definitions remains limited, and the proposed definitions are often unnatural and misalign with real scenarios, explicitly requiring models to ignore severe anomalies such as explosions or traffic accidents. In practice, events whose abnormality varies are more commonly minor rule violations, such as smoking or jaywalking. 
Consequently, this work proposes a new benchmark that introduces a more diverse set of natural anomaly definitions.

To address the above challenges, we propose \textsc{ESOM}, an \textbf{E}fficient \textbf{S}treaming \textbf{O}pen-world Video Anomaly Detection \textbf{M}odel, as illustrated in Fig.~\ref{fig:motivation}, which operates in a training-free manner and processes videos with a sliding-window strategy.
First, to tackle the open-world challenges, the model introduces a Definition Normalization (DN) module and a Probabilistic Scoring (PS) module. The DN module converts user prompt into a structured anomaly definition table to reduce hallucination, while the PS module converts MLLM's interval-level textual outputs into frame-level soft anomaly scores. 
Second, to enable efficient streaming inference, the model incorporates Inter-frame-matched Intra-frame Token Merging (IIM) and Hybrid Streaming Memory (HSM). Inspired by video coding, the IIM reduces visual tokens within each sliding window by dividing frames into groups with IBP-like references for spatiotemporal matching, enabling redundant tokens to be merged while preserving temporal localization. Meanwhile, the HSM manages streaming memory by reusing the KV cache from the previous window as short-term memory via inverse RoPE \cite{su2024roformer} rotation, and injecting prediction texts from earlier windows into prompts as long-term memory. 

For evaluation, this paper presents \textsc{OpenDef-Bench}, a high-quality benchmark comprising 770 videos and 1,492 samples, which provides diverse scene-dependent anomaly definitions and is built entirely from clean surveillance footage without camera cuts, visual effects, or watermarks. Unlike artificially constructed label partitions, its definitions reflect the natural context-dependent nature of anomalies, with definition shifts primarily arising from relatively subtle abnormal behaviors (e.g., smoking), thereby enabling more realistic evaluation.

Our contributions are summarized as follows:
\begin{enumerate}
    \item We propose ESOM, a training-free streaming framework for open-world video anomaly detection under dynamic anomaly definitions. It includes a Definition Normalization module for precise anomaly-definition conditioning and a Probabilistic Scoring module that converts interval-level textual outputs into frame-level soft anomaly scores.
    
    \item We design the IIM and HSM modules to improve efficiency, which reduce spatial-temporal redundancy for streaming videos and remain compatible with mainstream MLLMs and acceleration techniques.
    
    \item We introduce OpenDef-Bench, a new benchmark that provides diverse and natural dynamic anomaly definitions for evaluating open-world generalization.
    
    \item Extensive experiments demonstrate that ESOM achieves real-time inference on a single GPU while maintaining state-of-the-art performance in anomaly temporal localization, classification, and description generation.
\end{enumerate}

\section{Related Works}

\subsection{Open-world Video Anomaly Detection}

Recent advances in vision-language multimodal models have driven VAD toward a more open paradigm. Nevertheless, the concept of open-world VAD remains interpreted in different ways across the literature. 
One line of research \cite{Openset,huang2026openset,ubnormal,ovvad,anomize,cross-domain-generalization,cross-domain-limited,cross-domain-without} associates it with cross-domain generalization, where the primary goal is to handle unseen categories, novel scenes, and distribution shifts. 
Another line of work \cite{UCCD,bao2025anomaly_led,CUVA,UCA,HAWK,Holmes-VAU,wxx2025enhancing} emphasizes video anomaly understanding, suggesting that models should not only localize and classify anomalies, but also interpret and explain them in natural language. 
More recently, some studies \cite{cho2024multi-dataset,liu2025lagovad} have stressed the importance of modeling dynamically defined anomalies, requiring VAD systems to accurately detect abnormal events under varying user-provided definitions.

For cross-domain generalization, traditional methods often degrade substantially across domains. To address this issue, recent studies have explored open-set, open-vocabulary, and domain-generalization approaches: 
1) Openset: 
\cite{Openset} generates pseudo-anomalies in low-density regions via normalizing flows.
\cite{huang2026openset} constructs a human-prior feature space via semantic-aware transformation to learn domain-agnostic normality rules.
2) Open-vocabulary: 
OVVAD \cite{ovvad} is among the first to introduce vision-language models to match visual features with anomaly text descriptions for label assignment.
Anomize \cite{anomize} further improve the text branch with static and dynamic prompts to reduce detection ambiguity.
3) Domain-generalization: 
\cite{cross-domain-without} creates pseudo-abnormal examples in normal video frames to learn the relative difference between them.
However, these methods do not exploit the rich knowledge embedded in MLLMs, which limits their generalization ability.

For video anomaly understanding, recent studies formulate it as an MLLM-based QA problem \cite{HAWK,wxx2025enhancing}, which overlook the core task of temporal localization. 
A few methods attempt to localize anomalies alongside generating text, but they either provide only coarse segment-level predictions \cite{vadR1,vauR1} or rely on cascaded pipelines \cite{Holmes-VAU}. 
Therefore, a more integrated architecture is needed to jointly support fine-grained temporal localization and language-level anomaly understanding.

For dynamic anomaly definition, existing VAD methods can be broadly grouped by how anomalies are defined. Semi-supervised and unsupervised methods treat deviations from the normal data distribution as anomalies \cite{one-class-2017,one-class-wu2020}, while weakly supervised and fully supervised methods rely on dataset-provided category labels \cite{vadclip,GlanceVAD}. 
Despite their differences, both assume fixed anomaly definitions, meaning that an event considered abnormal during training cannot become normal at test time.
In contrast, \cite{liu2025lagovad} explicitly introduces dynamic anomaly definitions into VAD, but its understanding of anomaly semantics is still limited by model scale.
Although existing training-free MLLM-based methods can support dynamic detection via prompt engineering, they remain unexplored in this setting and exhibit unsatisfactory performance in our experiments.

However, these perspectives are largely studied separately. Existing methods usually address only one facet of open-world VAD, but fail to jointly support cross-domain generalization, language-level understanding with fine-grained localization, and dynamic anomaly definition. Therefore, a unified framework that integrates all three capabilities remains absent.

\subsection{VAD with Multimodal Large Language Models}
With the rapid development of MLLMs, an increasing number of studies have introduced them into video anomaly detection. Some methods draw inspiration from general video understanding and reformulate VAD as QA-style tasks. Among them, HAWK \cite{HAWK} focuses on motion-related anomaly cues, HolmesVAU \cite{Holmes-VAU} designs an anomaly-aware frame sampling module, and \cite{vadR1,vauR1,wxx2025enhancing} further incorporate chain-of-thought reasoning.
Another group of methods follows a training-free paradigm. LAVAD \cite{LAVAD} adopts a describe-then-score strategy, VERA \cite{vera} performs score estimation followed by post-processing, and PANDA \cite{PANDA} builds a more complex agent-based framework that improves performance through reflection and tool use.
Since intelligent surveillance is one of the primary application scenarios of VAD, streaming and real-time processing are of particular importance in practice. In such settings, the model should only access past observations and must keep pace with the incoming video stream. However, existing MLLM-based VAD methods largely overlook these practical requirements.

\subsection{Evaluation Methods for VAD}
Traditional VAD evaluation typically tests models on data drawn from the same distribution as the training set. Although recent methods achieve strong results under this setting, it is insufficient for assessing generalization.
To address this issue, some works \cite{Openset,huang2026openset,ubnormal,ovvad,anomize,cross-domain-generalization,cross-domain-limited,cross-domain-without} adopt cross-domain zero-shot evaluation, where models are tested on datasets with different distributions to examine robustness to unseen categories, novel scenes, and distribution shifts. However, these settings still assume static anomaly definitions.
To move beyond this limitation, \cite{liu2025lagovad} proposes Drift@5, a dynamic-definition evaluation protocol that constructs five anomaly definitions by selecting category subsets and reports the average performance across them. Nevertheless, the resulting definitions remain limited in diversity, and some are unnatural (\eg, treating explosions as normal events).
Therefore, we aim to build a larger-scale and more diverse benchmark for dynamic anomaly evaluation, where changes in anomaly definitions are restricted to relatively subtle or weak anomalies (e.g., smoking), leading to more realistic and meaningful evaluation scenarios.

\section{Method}

\begin{figure*}[t!]
    \centering
    \includegraphics[width=0.9\linewidth]{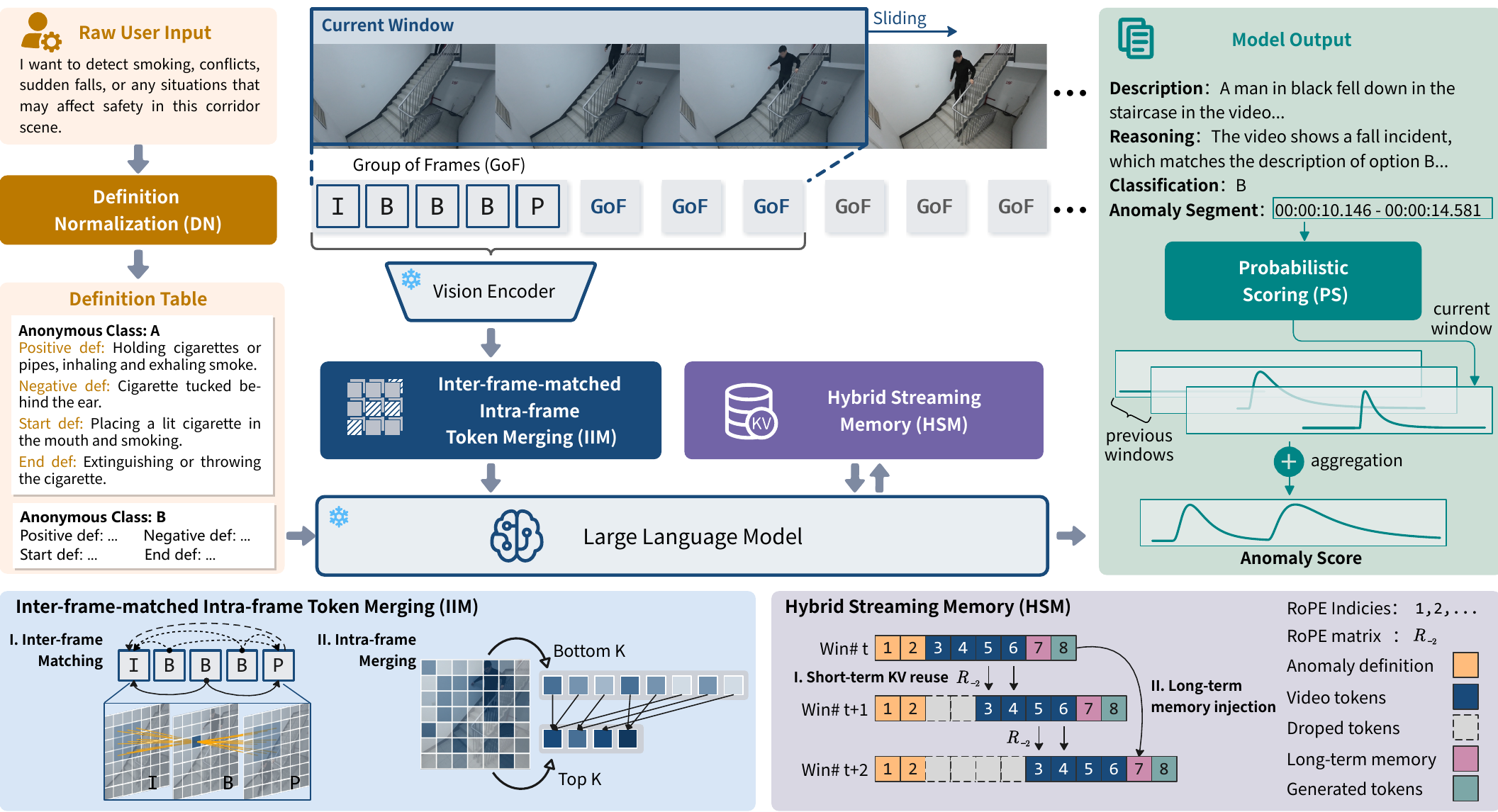}
    \caption{ESOM is a training-free streaming framework for open-world video anomaly detection under dynamic anomaly definitions. Given a raw user prompt, The DN module first converts it into a structured anomaly definition table. And the input video is processed in a sliding-window manner, where each window is encoded into visual tokens, and the IIM module compresses redundant tokens before they are fed into the LLM. 
    Meanwhile, the HSM module manages the model's cross-window memory states to maintain temporal continuity during streaming inference. 
    Based on the LLM outputs, including anomaly reasoning, classification, and temporal segments, the PS module further converts interval-level textual predictions into frame-level soft anomaly scores.}
    \label{fig:method}
\end{figure*}

\subsection{Problem Formulation}
\label{sec:formulation}

Given an input video $V=\{I_m\}_{m=0}^{T-1}$ with $T$ frames and a anomaly definition $z$ provided by the user, where $z$ specifies a category set $\mathcal{C}=\{c_1,\dots,c_K\}$, the task of open-world video anomaly detection discussed in this paper is to produce three types of outputs: a frame-level anomaly score sequence $S=\{S(m)\}_{m=0}^{T-1}$ with $S(m)\in[0,1]$, a generated textual prediction sequence $\{y_t\}$ over sliding windows, and the corresponding predicted category sequence $\{c_t\}$ parsed from the generated texts, where $c_t \in \mathcal{C}$. 

\subsection{Framework Overview}

As shown in Fig.~\ref{fig:method}, ESOM is a training-free streaming framework for open-world video anomaly detection under dynamic anomaly definitions. Given a user-defined anomaly description and an input video stream, the framework produces textual explanations, category labels, and frame-level anomaly scores.
For each sliding window, the DN module first converts the raw user prompt into a structured definition. The video frames are then encoded into visual tokens and compressed by the IIM module. Based on the normalized definition and compressed tokens, the MLLM performs inferencing autoregressively, while the HSM module manages the memory across the windows. Finally, the PS module converts interval-level predictions into frame-level anomaly scores.
These modules together enable efficient streaming inference and adaptation to dynamic anomaly definitions.
The details of these modules are as follows.

\subsection{Definition Normalization}
\label{sec:def_norm}

In open-world video anomaly detection, anomaly definitions are often provided in free-form natural language, which can be ambiguous for understanding~\cite{liu2025lagovad}. To improve semantic clarity, a Definition Normalization (DN) module converts the raw user definition $z$ into a structured definition table through an external LLM:
\begin{equation}
    z^{n} = \mathcal{N}(z, P_{\mathrm{norm}}),
\end{equation}
where $\mathcal{N}(\cdot)$ denotes the normalization function, $P_{\mathrm{norm}}$ is the prompt, and $z^{n}$ is the normalized definition.

The normalized definition includes anonymous class options, positive definitions, negative definitions, and event boundary definitions for each category. 
Anonymous class options replace original category names with neutral identifiers (\eg, letters) to reduce semantic bias. 
Positive definitions describe what events should be regarded as belonging to each category, while negative definitions specify confusing but excluded cases. 
Event boundary definitions further clarify when an anomalous event should be considered to start and end, thereby standardizing the temporal criteria for anomaly localization. 
Since DN is only triggered when the anomaly definition changes, it can be handled by a black-box LLM with no extra overhead during streaming inference.

\subsection{Inter-frame-matched Intra-frame Token Merging}
\label{sec:IIM}

During inference, visual tokens dominate the computational cost and are therefore the primary target for efficiency optimization. To design a visual token compression module, it should first remain training-free and compatible with general-purpose MLLMs, so that it can be readily adapted to new models without finetuning. Second, it should support streaming inference by performing token compression stably within each window. Finally, it should preserve temporal information, since accurate anomaly localization depends on fine-grained temporal cues. Existing strategies such as keyframe selection \cite{tang2025aks} or inter-frame feature fusion \cite{framefusion} are less suitable in our setting because they either change the temporal sampling density or blur temporal positions.

Inspired by the Group-of-Frames (GoF) design in video coding (\eg, H.264~\cite{h264}), each input window is first divided into multiple GoFs, and the frames within each GoF are further assigned as I-, P-, and B-frames. The I-frame is kept intact, while the P- and B-frames are matched with their reference frames to identify redundant patches, which are then compressed by intra-frame token merging. In this way, the temporal structure required for streaming localization is preserved.
The specific process is divided into two steps: 

\paragraph{Inter-frame matching}

For the current window $t$, the sampled frames are encoded into visual features $v_t$, which are further divided into GoFs of size $l_g$. Omitting the GoF index for brevity, let $g_j^{u,v} \in \mathbb{R}^d$ denote the feature of the patch at spatial position $(u,v)$ in frame $j$ of a GoF.
For each frame, reference frames are selected according to its frame type. For a P-frame, the reference is the previous I-frame or P-frame; for a B-frame, the references are the preceding and following I/P-frames. Let $R(j)$ denote the index set of reference frames for frame $j$. For any patch $(u,v)$, its importance score is defined as the minimum cosine distance to the local neighborhood $N(u,v)$ in the reference frames:
\begin{equation}
o_j^{u,v}
=
\min_{r \in R(j)} \ \min_{(u',v') \in N(u,v)}
\left(1 - \cos\left(g_j^{u,v}, g_r^{u',v'}\right)\right).
\end{equation}
A smaller value of $o_j^{u,v}$ indicates stronger redundancy.

\paragraph{Intra-frame merging}
Given the retained proportion $\gamma$, the patches of frame $j$ are ranked according to their importance scores $\{o_j^{u,v}\}$. The top $\lfloor \gamma hw \rfloor$ patches form the high-score set $H(j)$, and the remaining patches form the low-score set $L(j)$. 
Inspired by the I/B/P frame design in video coding, the retained proportion is set differently for different frame types, with $\gamma_B < \gamma_P < \gamma_I = 1$.

For each low-score patch $(u,v) \in L(j)$, its most similar target patch in the high-score set $H(j)$ is defined as
\begin{equation}
\pi_j(u,v)
=
\operatorname*{arg\,max}_{(p,q)\in H(j)}
\cos\left(g_j^{u,v}, g_j^{p,q}\right).
\end{equation}
Accordingly, the low-score patches assigned to a high-score patch $(p,q) \in H(j)$ form the set $\mathcal{A}_j^{p,q}=\left\{(u,v)\in L(j)\ \middle|\ \pi_j(u,v)=(p,q)\right\}$
Then, each high-score patch $(p,q) \in H(j)$ is updated by merging the patches in $\mathcal{A}_j^{p,q}$ with similarity-based weights:
\begin{equation}
\begin{aligned}
\alpha_{(u,v)\rightarrow(p,q)}
&=
\frac{
\exp\left(\cos\left(g_j^{u,v}, g_j^{p,q}\right)\right)
}{
\sum_{(u',v')\in \mathcal{A}_j^{p,q}}
\exp\left(\cos\left(g_j^{u',v'}, g_j^{p,q}\right)\right)
}, \\
\tilde{g}_j^{p,q}
&=
\frac{1}{2} g_j^{p,q}
+
\frac{1}{2} \sum_{(u,v)\in \mathcal{A}_j^{p,q}}
\alpha_{(u,v)\rightarrow(p,q)} \, g_j^{u,v}.
\end{aligned}
\end{equation}
After merging, only the updated high-score tokens $\{\tilde{g}_j^{p,q}\}_{(p,q)\in H(j)}$ are retained for each frame. The compressed visual representation of the current window, denoted by $\tilde{v}_t$, is obtained by concatenating all retained tokens in their original temporal and spatial order.

\subsection{Hybrid Streaming Memory}
\label{sec:HSM}

When processing streaming videos, preserving all historical information is infeasible under limited memory budgets, while temporal windows at different positions also contribute unequally.
Therefore, a Hybrid Streaming Memory (HSM) module is introduced to combine short-term KV reuse with compact long-term memory injection. 
On the one hand, the short-term memory in the overlap between windows $t-1$ and $t$ is reused to eliminate redundant computation. However, as shown in Fig.~\ref{fig:method}, this reuse disrupts the continuity of temporal encoding between anomaly-definition tokens and video tokens. Inspired by~\cite{xiao2023streamingllm}, a reverse RoPE matrix is applied to the reused KV cache to restore temporally consistent encoding. 
On the other hand, the outputs from windows earlier than $t-1$ are summarized into long-term memory and injected into the current window in textual form at low cost.

\paragraph{Short-term KV reuse}
Specifically, the input sequence of the MLLM for the $t$-th window consists of four parts:
\begin{equation}
x_t = \mathrm{Concat}[P^{\mathrm{sys}}, s^n, \tilde{v}_t, P_{t-1}^{\mathrm{mem}}],
\end{equation}
where $P^{\mathrm{sys}}$ denotes the system prompt used to constrain the task behavior and output format of the model, $s^n$ is the normalized prompt text obtained in Sec.~\ref{sec:def_norm}, $\tilde{v}_t$ is the compressed visual feature obtained in Sec.~\ref{sec:IIM}, and $P_{t-1}^{\mathrm{mem}}$ is the long-term memory text from previous windows.

Since $P^{\mathrm{sys}}$ and $s^n$ remain unchanged throughout inference until the anomaly definition changes, their shared prefix KV cache, denoted by $(K^{\mathrm{pre}}, V^{\mathrm{pre}})$, is computed only once:
\begin{equation}
K^{\mathrm{pre}}, V^{\mathrm{pre}}
=
\mathcal{M}_{\mathrm{prefill}}
\bigl(
K^0, V^0, \mathrm{Concat}[P^{\mathrm{sys}}, s^n]
\bigr),
\end{equation}
where $\mathcal{M}_{\mathrm{prefill}}(K, V, x)$ denotes the prefilling stage of the MLLM that appends the token sequence $x$ to the existing KV cache and outputs the updated cache, and $K^0, V^0$ is the empty cache.

For the visual input, the visual KV cache of the first window, $(K_{1}^{\mathrm{vis}}, V_{1}^{\mathrm{vis}})$, is computed through a standard prefilling:
\begin{equation}
K_{1}^{\mathrm{vis}}, V_{1}^{\mathrm{vis}}
=
\mathcal{M}_{\mathrm{prefill}}
\bigl(
K^{\mathrm{pre}}, V^{\mathrm{pre}}, \tilde{v}_{1}
\bigr).
\end{equation}
For subsequent windows, we exploit the relative property of RoPE~\cite{su2024roformer} and apply an inverse rotation to the overlapped key cache, which is equivalent to shifting its RoPE indices backward by $M'$. For window $t>1$, the contextual KV cache $(K_t^{\mathrm{ctx}}, \, V_t^{\mathrm{ctx}})$ is computed as
\begin{equation}
\begin{aligned}
K_t^{\mathrm{ctx}}, \, V_t^{\mathrm{ctx}}
= \mathcal{M}_{\mathrm{prefill}} \Big(
&\mathrm{Concat}[K^{\mathrm{pre}}, R_{-M'} K_{t-1}^{\mathrm{ov}}], \\
&\mathrm{Concat}[V^{\mathrm{pre}}, V_{t-1}^{\mathrm{ov}}], \tilde{v}_t^{\mathrm{new}}
\Big),
\end{aligned}
\end{equation}
where $(K_{t-1}^{\mathrm{ov}}, V_{t-1}^{\mathrm{ov}})$ denote the overlapping tail of the previous visual KV cache, $\tilde{v}_t^{\mathrm{new}}$ denotes the newly added visual tokens in the current window, and $R_{-M'}$ denotes the corresponding rotation matrix.

\paragraph{Long-term memory injection}
The long-term memory is updated autoregressively across windows. For the current window $t$, the model obtains the full cache $(K_t^{\mathrm{full}}, V_t^{\mathrm{full}})$ by appending the previous memory text $P_{t-1}^{\mathrm{mem}}$:
\begin{equation}
K_t^{\mathrm{full}}, V_t^{\mathrm{full}} = \mathcal{M}_{\mathrm{prefill}}\bigl(K_t^{\mathrm{ctx}}, V_t^{\mathrm{ctx}}, P_{t-1}^{\mathrm{mem}}\bigr).
\end{equation}
The model then generates the prediction $y_t$ with the full cache and updates the memory for the next window:
\begin{equation}
y_t = \mathcal{M}_{\mathrm{dec}}\big(K_t^{\mathrm{full}}, V_t^{\mathrm{full}}\big), \quad
P_t^{\mathrm{mem}} = \mathcal{F}_{\mathrm{mem}}\big(P_{t-1}^{\mathrm{mem}}, y_t\big)
\end{equation}
where $\mathcal{M}_{\mathrm{dec}}(K, V)$ denotes the decode stage of the MLLM, $\mathcal{F}_{\mathrm{mem}}(\cdot)$ denotes a memory update function that retains the generated descriptions of the most recent $M$ windows, maintaining compact long-range context while minimizing token cost. Finally, the predicted category is parsed from the generated text $c_t = \mathcal{F}_{\mathrm{parse}}(y_t)$, where the parse function maps the generated text to a category in $\mathcal{C}$.

\subsection{Probabilistic Scoring}
\label{sec:PS}

The MLLM produces temporal localization results in textual form, which cannot be directly used for frame-level evaluation. To bridge this gap, a Probabilistic Scoring (PS) module is introduced to transform each predicted interval into a local probability curve and then aggregate the curves from overlapping sliding windows into a global frame-level anomaly score sequence. 
The module is implemented in two steps.

\paragraph{Interval probabilization}
First, we probabilize the interval output along the temporal dimension.
For each window $t$, the parsed temporal interval is $[\tau_t^s, \tau_t^e]$. Let the window length be $l$ frames, and let the local frame index be $x \in \{0,1,\dots,l-1\}$. A log-normal kernel is then defined as
\begin{equation}
k_t(x)=
\begin{cases}
\dfrac{1}{x\sigma_k}\exp\!\left(-\dfrac{(\ln x-\mu_t)^2}{2\sigma_k^2}\right), & x>0,\\[6pt]
0, & x\leq 0,
\end{cases}
\end{equation}
where $\sigma_k$ is the shape parameter, $\mu_t$ is the location parameter, and $k_t(x)$ is set to zero if no valid interval is parsed. 
To place the peak at relative position $r \in (0,1)$ within the predicted interval, we set $\mu_t = \ln\!\bigl((\tau_t^e-\tau_t^s)r+\tau_t^s\bigr) + \sigma_k^2$.
In our implementation, we set $r=0.2$ to impose a prior that anomaly onsets are typically easier to identify consistently than anomaly endings due to the subjective nature of anomaly annotation, as discussed in \cite{liu2025rethinking}.

\paragraph{Confidence calibration}
Next, we also probabilize the output at the window level.
Specifically, we use the average difference between the top-1 and top-2 token probabilities over all generated tokens in the current window \cite{wang2024chain} as the window-level confidence $\alpha_t$, and smooth the resulting score with a Gaussian kernel $g_{\sigma_g}$ with standard deviation $\sigma_g$:
\begin{equation}
a_t(x)=\alpha_t \, (g_{\sigma_g} * k_t)(x).
\end{equation}
After obtaining the local anomaly scores, we accumulate the scores from all windows to produce the global frame-level anomaly score sequence:
\begin{equation}
A_t(m)=
\begin{cases}
a_t(m-s_t), & m \in [s_t, s_t+l-1],\\
0, & \text{otherwise}.
\end{cases}
\end{equation}
\begin{equation}
S(m)=\dfrac{\sum_t A_t(m)}{\max_m \sum_t A_t(m)},
\end{equation}
where $m \in \{0,1,\dots,T-1\}$ is the global frame index, and $s_t$ denotes the starting frame index of window $t$ on the global timeline.
Under a fixed window length and overlap ratio, the normalization factor $\max_m \sum_t A_t(m)$ can be determined in advance, making the entire scoring process compatible with streaming inference.

\section{Benchmark: OpenDef-Bench}

To comprehensively evaluate model performance for video anomaly detection under open-world scenarios with dynamic anomaly definitions, we propose \textbf{OpenDef-Bench}, an \textbf{Open}-world dynamic \textbf{def}inition video anomaly detection \textbf{bench}mark. 
For each video, OpenDef-Bench provides multiple anomaly definitions, each containing several candidate categories. All definitions are manually verified to ensure that they are natural and consistent with the scene context. For example, OpenDef-Bench includes videos of pets entering meeting rooms, for which both positive (\ie, treating it as abnormal) and negative definitions (\ie, treating it as normal) are provided. In contrast, the existing dynamic definition method \cite{liu2025lagovad} directly treats events such as car accidents or fights as either abnormal or normal, while regarding such events as normal is generally unnatural and may reduce the realism of the evaluation setting.
We introduce the construction pipeline, evaluation tasks and statistics of OpenDef-Bench as follows.

\subsection{Construction Pipeline}

Fig.~\ref{fig:benchmark}(a) illustrates the four steps of the pipeline:

\begin{figure}[!t]
    \centering
    \includegraphics[width=0.99\linewidth]{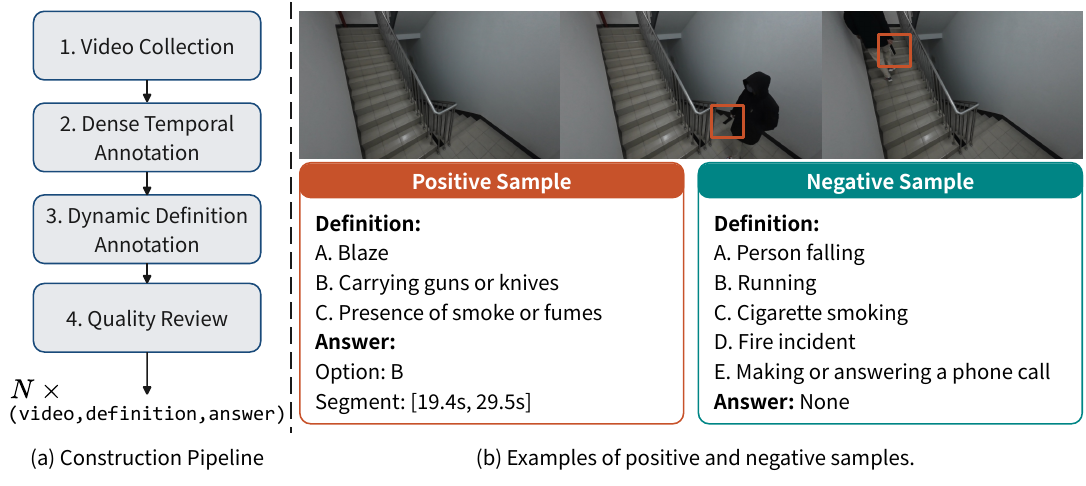}
    \caption{The construction pipeline and example samples of the proposed OpenDef-Bench. Each video is annotated with multiple definitions containing multiple candidate categories. A sample is a triplet of a video, a definition, and the corresponding answer.}
    \label{fig:benchmark}
\end{figure}

\begin{enumerate}
    \item \textbf{Video collection.}
    OpenDef-Bench targets surveillance and long-video scenarios. It is constructed from multiple sources, including public datasets such as ShanghaiTech~\cite{ShanghaiTech}, NWPU~\cite{NWPU}, VIRAT~\cite{virat}, and MEVA~\cite{meva}, together with our newly recorded videos. During collection, watermarks, visual effects, and editing artifacts are strictly avoided. To ensure video diversity, the data is balanced across factors such as day and night, indoor and outdoor scenes, and target distance.

    \item \textbf{Dense temporal annotation.}
    Annotators densely label all events in each video with frame-level temporal boundaries that may be considered abnormal under certain definitions. This step is designed to cover events whose abnormality depends on the definition.

    \item \textbf{Dynamic definition annotation.}
    Annotators provide multiple definitions for the events obtained from the previous step. They first review the full video and then propose multiple scene-appropriate anomaly categories. Each event is subsequently annotated as a positive or negative sample under different definitions. For positive samples, 3--20 candidate categories are randomly selected, with exactly one matching the event; for negative samples, all selected categories are irrelevant. To further increase the difficulty, categories are allowed to be free-form phrases, and some of the categories are designed to require fine-grained discrimination, such as ``riding a bicycle with one hand'' and ``cycling with a passenger''.

    \item \textbf{Resample and review.}
    Finally, the collected data is resampled to balance the distribution across scenes and categories, while a cross-review process is conducted to remove samples with inaccurate temporal boundaries or ambiguous semantics. Each sample is organized as a ``video--definition--answer'' triplet.
\end{enumerate}

\subsection{Evaluation Tasks}

OpenDef-Bench supports two evaluation tasks: \textbf{dynamic temporal localization} and \textbf{dynamic category selection}:
For dynamic temporal localization, the model is required to produce frame-level anomaly scores conditioned on dynamic anomaly definitions. These scores are then evaluated using temporal localization metrics based on the frame-level annotations provided by the benchmark.
For dynamic category selection, the model is required to select the correct category from the dynamically provided anomaly definitions, or predict normal if none of the defined categories is present. This task is evaluated using accuracy, which measures the model’s ability to perform fine-grained category discrimination under dynamically specified anomaly definitions.

\begin{table}[t]
\centering
\caption{Comparison of OpenDef-Bench with existing VAD benchmarks. Clean: w/o watermarks, editing, or visual effects.}
\label{tab:benchmark_comparison}
\setlength{\tabcolsep}{3pt}
\resizebox{\columnwidth}{!}{%
\begin{tabular}{lcccccc}
\toprule
\textbf{Benchmark} & \textbf{Dynamic} & \textbf{Clean} & \textbf{\# Videos} & \textbf{\# Samples} & \textbf{Duration (h)} & \textbf{Resolution} \\
 & \textbf{Definitions} &  &  &  &  &  \\
\midrule
UCF-Crime~\cite{ucf-crime} 
& $\times$ & $\times$ 
& 290 & 290 & 10.3 & 240P \\
XD-Violence~\cite{xdviolence} 
& $\times$ & $\times$ 
& \textbf{800} & 800 & 27.0 & 360P \\
MSAD~\cite{msad} 
& $\times$ & $\times$ 
& 240 & 240 & 1.4 & 720P, 1080P \\
ShanghaiTech~\cite{ShanghaiTech} 
& $\times$ & $\checkmark$ 
& 107 & 107 & 0.5 & 480P \\
NWPU~\cite{NWPU} 
& $\times$ & $\checkmark$ 
& 242 & 242 & 4.3 & 1080P \\
\textbf{OpenDef-Bench} 
& $\checkmark$ & $\checkmark$ 
& 770 & \textbf{1492} & \textbf{27.7} & \textbf{480P, 1080P} \\
\bottomrule
\end{tabular}
}
\vspace{-2em}
\end{table}

\subsection{Data Statistics}

Table~\ref{tab:benchmark_comparison} compares OpenDef-Bench with representative video anomaly detection benchmarks. Existing benchmarks such as UCF-Crime~\cite{ucf-crime}, XD-Violence~\cite{xdviolence}, and MSAD~\cite{msad} cover diverse scenarios, but they use fixed anomaly definitions and do not provide dynamic definition annotations. In addition, they mainly collect data from online videos, which often contain heavy watermarks and editing artifacts. Datasets such as ShanghaiTech~\cite{ShanghaiTech} and NWPU~\cite{NWPU} rely on manually recorded surveillance videos and therefore better reflect real surveillance environments. However, they remain limited in both scale and scenario diversity, and they do not support the dynamic definition required in open-world settings.

\begin{figure}[!t]
    \centering
    \captionsetup[subfloat]{font=scriptsize}

    \subfloat[{Resolution Distribution\label{fig:stat_res}}]{
        \includegraphics[width=0.4\columnwidth]{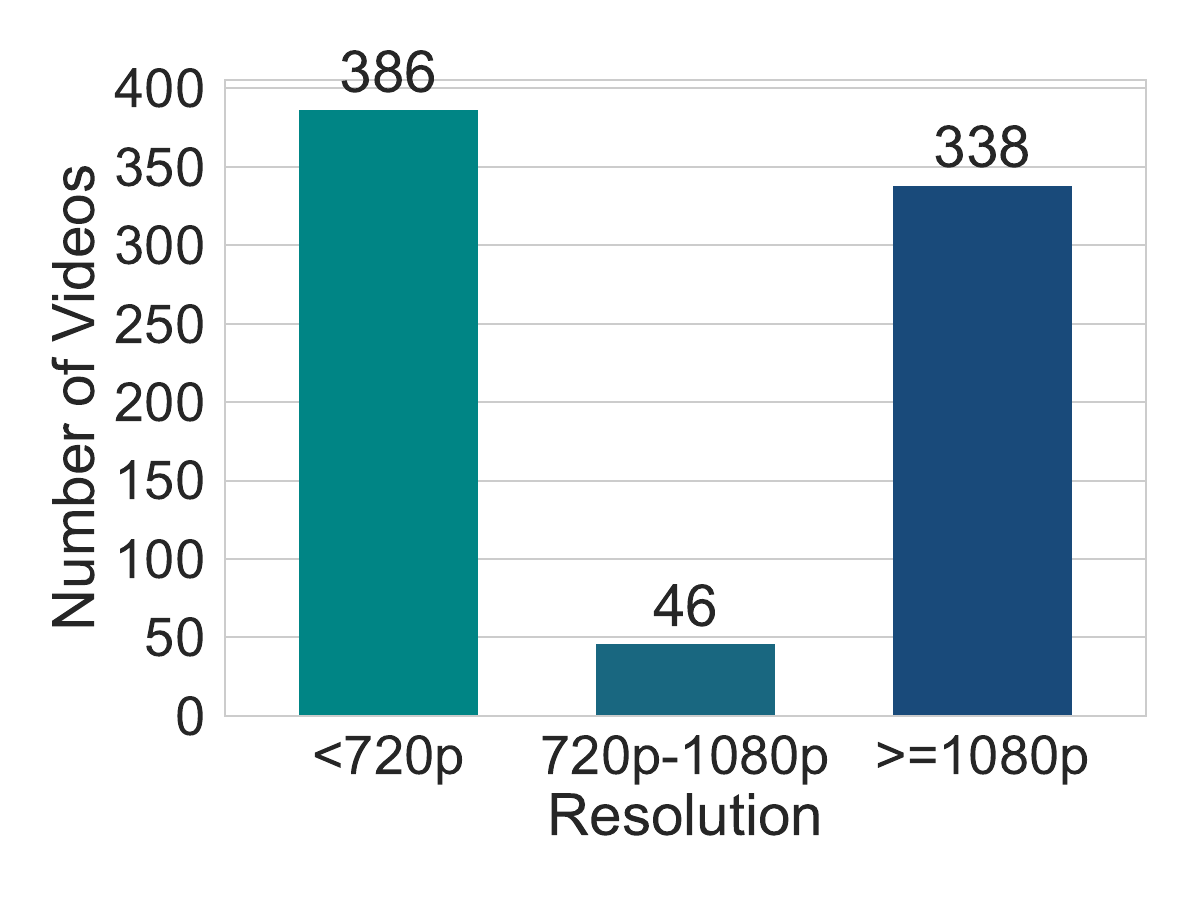}
    }
    \hspace{0.04\columnwidth}
    \subfloat[{Duration Distribution\label{fig:stat_dur}}]{
        \includegraphics[width=0.4\columnwidth]{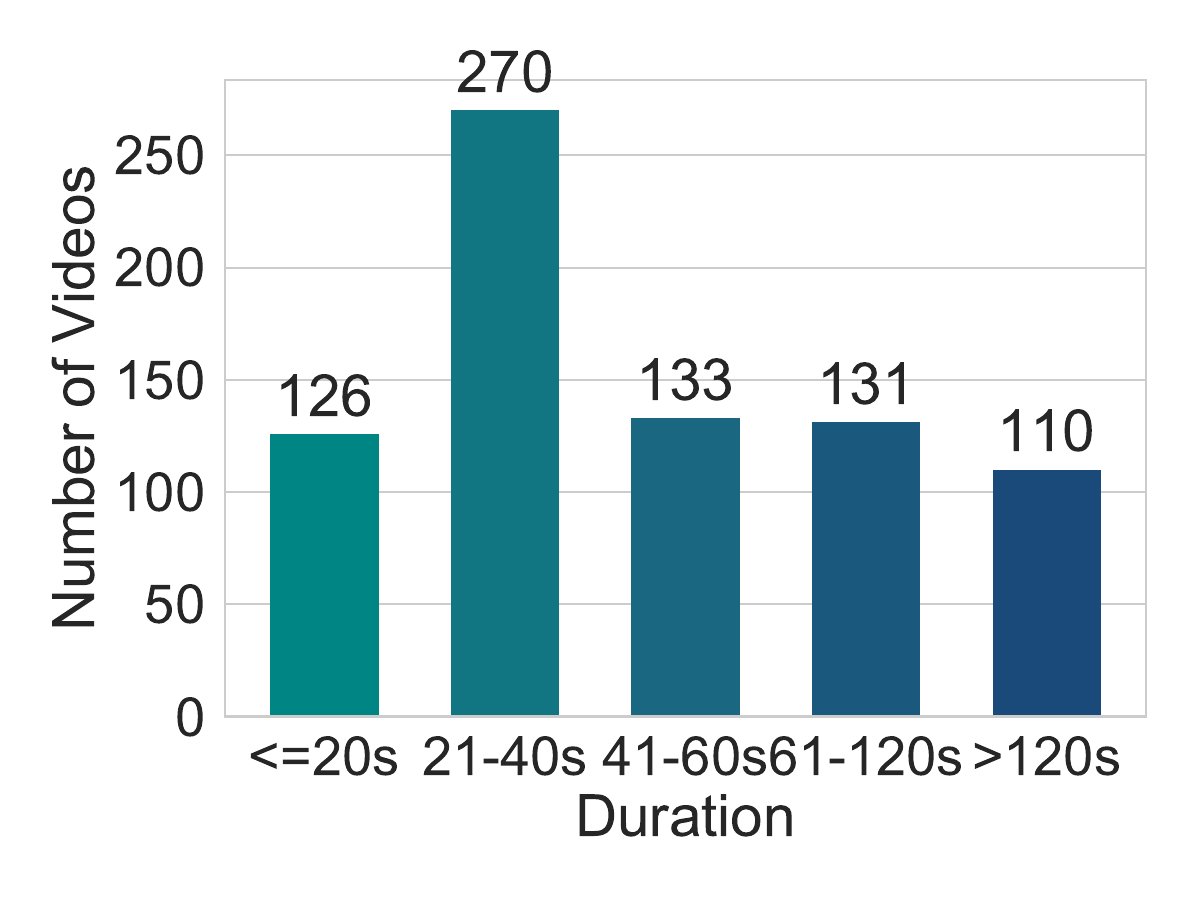}
    }

    \vspace{-1em}

    \subfloat[Option Distribution\label{fig:stat_opt}]{
        \includegraphics[width=0.4\columnwidth]{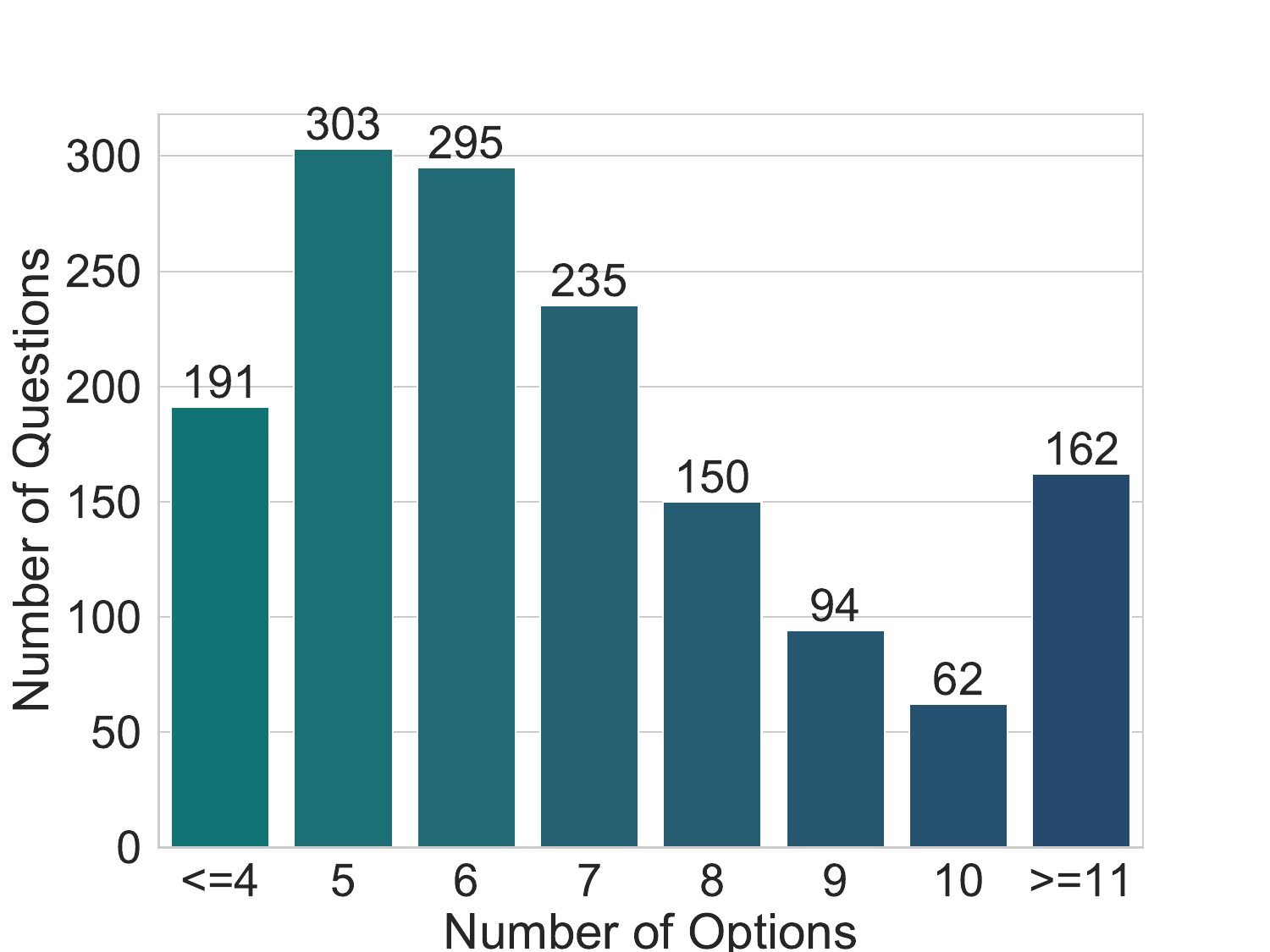}
    }
    \hspace{0.06\columnwidth}
    \subfloat[Answer Distribution.\label{fig:stat_ans}]{
        \includegraphics[width=0.3\columnwidth]{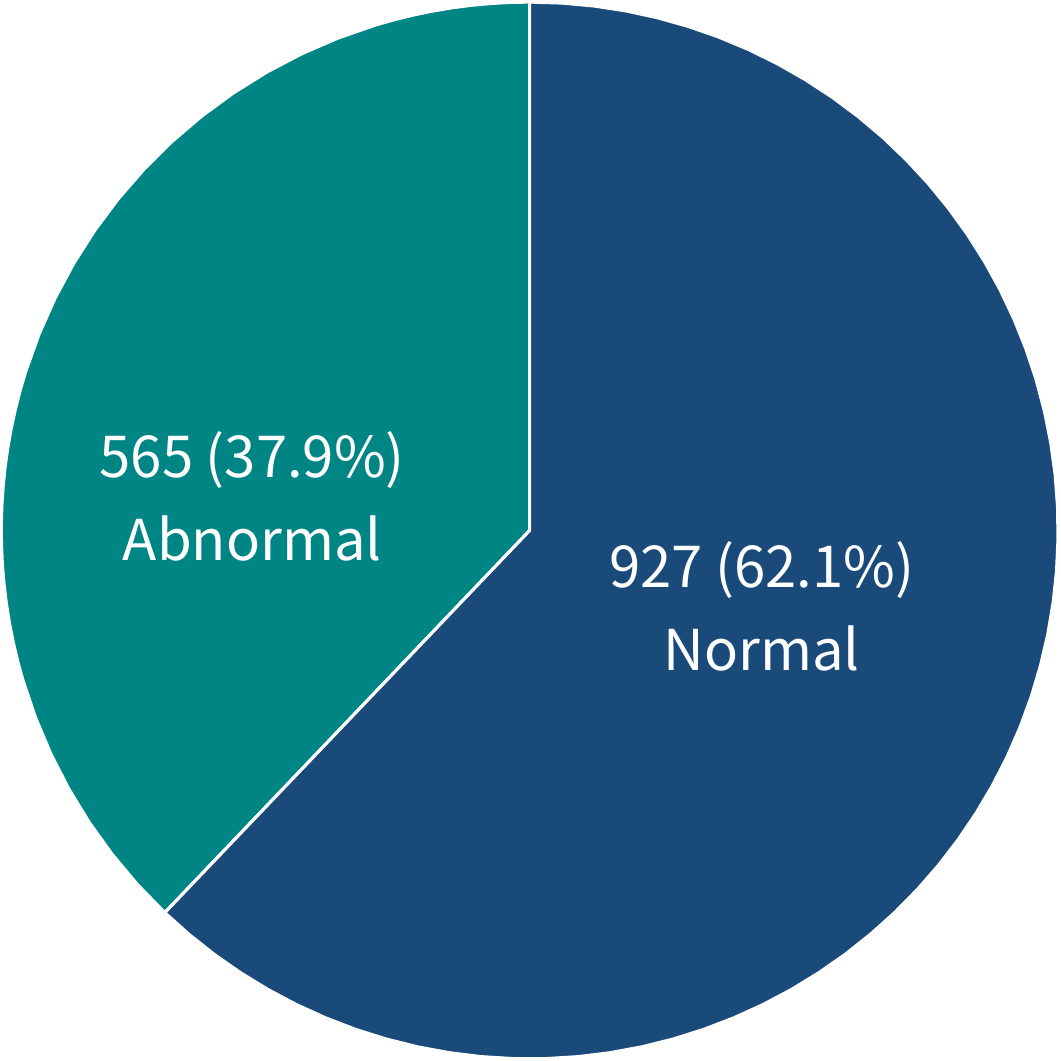}
    }

    \caption{Statistics of OpenDef-Bench. With high video resolution, long video duration, and diverse answer options, OpenDef-Bench presents a challenging benchmark for evaluation.}
    \label{fig:stat}
\end{figure}

By contrast, as shown in Table~\ref{tab:benchmark_comparison} and Fig.~\ref{fig:stat}, OpenDef-Bench includes videos without watermarks and editing while introducing dynamic definitions. It also covers more videos, more samples, and longer video duration, thus providing a more reliable and challenging benchmark for evaluating model generalization under dynamic anomaly definitions.

\begin{figure}
    \centering
    \includegraphics[width=0.85\linewidth]{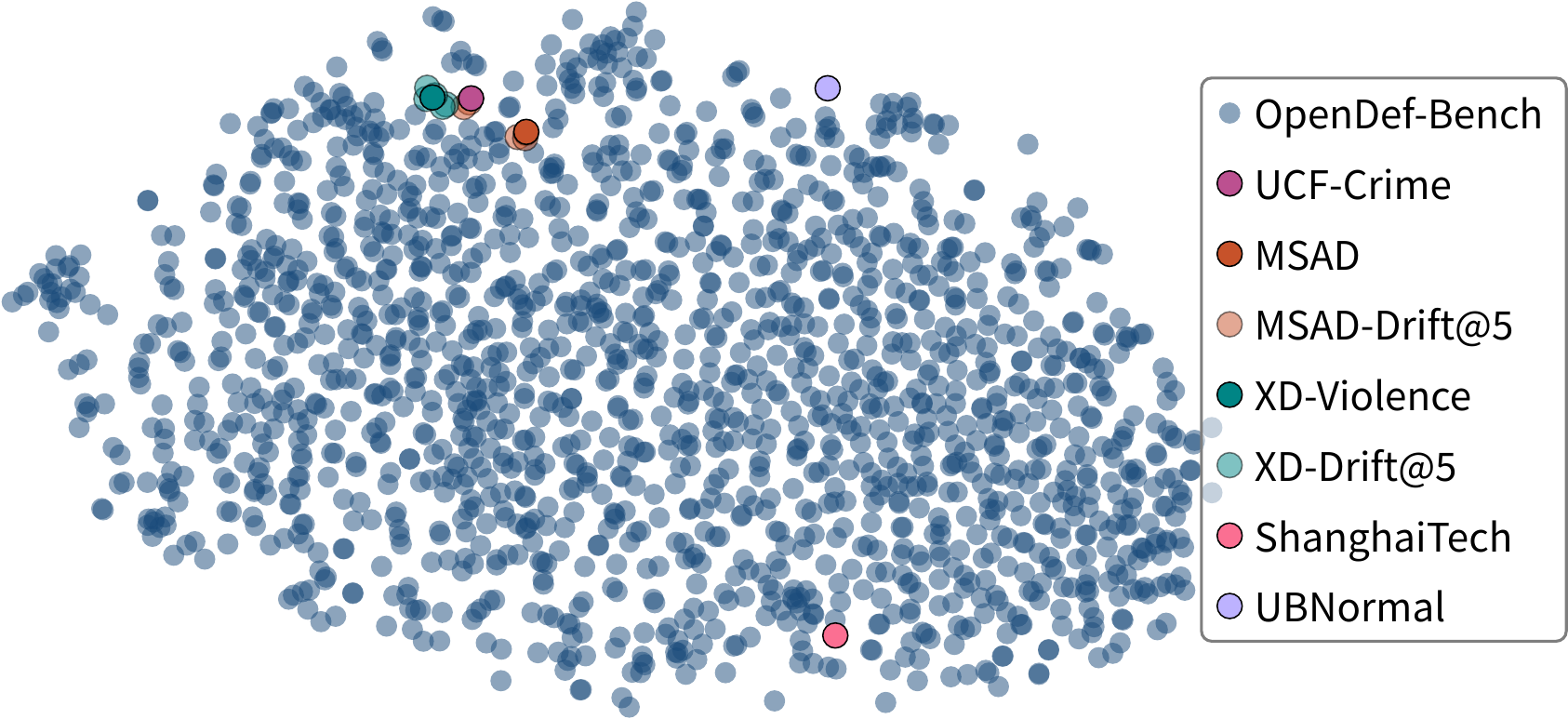}
    \caption{T-SNE visualization of text embeddings of anomaly definitions from OpenDef-Bench and existing VAD benchmarks. Definitions from existing benchmarks cluster in localized regions, while those from OpenDef-Bench are more uniformly distributed across the semantic space.}
    \label{fig:tsne}
    \vspace{-1em}
\end{figure}

To verify the diversity of anomaly definitions, we concatenate the category texts and extract their text embeddings using Qwen3 \cite{yang2025qwen3}, followed by T-SNE visualization. As shown in Fig.~\ref{fig:tsne}, traditional benchmarks adopt a single fixed anomaly definition, which corresponds to only one point in the visualization. Drift@5 \cite{liu2025lagovad} introduces five anomaly definitions and therefore corresponds to five points. However, we observe that these definitions are highly semantically similar, making them insufficient for effectively evaluating definition diversity. In contrast, OpenDef-Bench covers a much broader range of anomaly definitions, enabling a more comprehensive evaluation of model performance under dynamic-definition scenarios.
More examples and a dataset statement are provided in the supplementary material.

\section{Experiments}

\subsection{Evaluation Methods}

We compare the proposed method with existing approaches from four perspectives in the open-world setting.

\begin{itemize}
    \item \textbf{Traditional zero-shot performance} is evaluated through localization and classification tasks in the zero-shot setting on UCF-Crime \cite{ucf-crime}, XD-Violence \cite{xdviolence}, and ShanghaiTech \cite{ShanghaiTech}. Specifically, for ShanghaiTech, we adopt the split \cite{PEL} that includes both normal and abnormal videos.
    For temporal localization, we use AUC and AP metrics, together with the LaAP metric proposed in~\cite{liu2025rethinking}, which measures whether a model can detect anomalies at an early stage of events. For classification, we use Macro-F1 to mitigate class imbalance.
    
    \item \textbf{Dynamic definition performance} is evaluated with two tasks on OpenDef-Bench and the Drift@5 \cite{liu2025lagovad} on XD-Violence. For the dynamic temporal localization task, we adopt the AUC, AP and LaAP. For the dynamic category selection task, we use Accuracy. 

    \item \textbf{Explainability}. We evaluate the quality of generated descriptions on HIVAU-70K \cite{Holmes-VAU}. Specifically, due to the lack of dedicated datasets for streaming inference explainability, we use 398 video-level annotated samples from this dataset and employ GPT-4o as a judge to compare the generated descriptions with those produced by HolmesVAU \cite{Holmes-VAU} fine-tuned on HIVAU-70K. The judge analyzes content accuracy, completeness, and hallucination, and reports the win rate (WR).

    \item \textbf{Efficiency}. We compare efficiency under the same hardware platform and environment using the Real-Time Factor (RTF) \cite{sun2025streamavatar_RTF}, defined as the ratio between model processing time and input video duration. When RTF is less than 1, the model can process video streams in real time. We report the average RTF over multiple long videos as the final result.
\end{itemize}

\subsection{Implementation Details}

We use the non-thinking version of \texttt{Qwen3-VL-8B} as default. 
Unless otherwise specified, all experiments are conducted on a single GPU with 48G memory, implemented with \texttt{flash-attention 2} and \texttt{bfloat16}, while no quantization is applied.
For streaming video inference, a sampling rate of 4 fps is used, with a sliding window length of 224 frames and an overlap of 160 frames between adjacent windows.
For the IIM module, a GoF structure with size 8 is adopted, where the first frame is treated as an I-frame, the last frame as a P-frame, and the remaining frames as B-frames.
Correspondingly, the token retention ratios for B-frames and P-frames are set to $\gamma_B = 0.2$ and $\gamma_P = 0.6$, respectively.
In the PS module, the relevant hyperparameters are set as $r = 0.2$, $\sigma_k = 1.2$, and $\sigma_g = 4$.

\subsection{Comparison with state of the art}

\begin{table}[t]
\centering
\small
\setlength{\tabcolsep}{3pt}
\caption{Comparison with existing methods on traditional zero-shot video anomaly detection benchmarks. T denotes Training, GF denotes Gradient-free, and TF denotes Training-free.}
\label{tab:exp_zeroshot}
\resizebox{\columnwidth}{!}{%
\begin{tabular}{lccccc ccc cc}
\toprule
\multirow{2}{*}{Model}
& \multirow{2}{*}{\shortstack{Type}}
& \multirow{2}{*}{\shortstack{Online}}
& \multicolumn{3}{c}{UCF-Crime}
& \multicolumn{3}{c}{XD-Violence}
& \multicolumn{2}{c}{ShanghaiTech} \\
\cmidrule(lr){4-6} \cmidrule(lr){7-9} \cmidrule(lr){10-11}
& & 
& AUC & LaAP & F1
& AP & LaAP & F1
& AUC & LaAP \\
\midrule

VadCLIP~\cite{vadclip} & T & $\times$ 
& 80.16 & 21.70 & 10.52
& 58.29 & 60.16 & 26.16
& 60.73 & 9.20 \\

LaGoVAD~\cite{liu2025lagovad} & T & $\times$
& 81.11 & 22.48 & 16.64
& 74.25 & 75.82 & 63.80
& 45.74 & 4.94 \\

Qwen3-VL-8B & TF & $\times$
& 63.64 & 6.33 & 21.79
& 44.36 & 33.92 & 69.02
& 58.68 & 4.38 \\

LAVAD~\cite{LAVAD} & TF & $\times$
& 79.21 & 24.01 & --
& -- & -- & --
& 57.58 & 6.52 \\

VERA~\cite{vera} & GF & $\times$
& 85.57 & 29.82 & --
& 56.24 & 56.61 & --
& 59.10 & 16.36 \\
\midrule

VERA & GF & $\checkmark$
& 75.27 & 16.87 & --
& 44.22 & 50.97 & --
& 54.85 & 6.48 \\

PANDA~\cite{PANDA} & TF & $\checkmark$
& 82.57 & -- & --
& \textbf{72.41} & -- & --
& -- & -- \\

Qwen3-VL-8B & TF & $\checkmark$
& 71.99 & 18.06 & 35.63
& 29.94 & 27.55 & 19.23
& 56.20 & 3.85 \\

\textbf{ESOM} & TF & $\checkmark$
& \textbf{86.18} & \textbf{33.21} & \textbf{41.26}
& 71.68 & \textbf{73.64} & \textbf{73.57}
& \textbf{77.06} & \textbf{34.33} \\
\bottomrule
\end{tabular}
}
\end{table}

\begin{table}[t]
\centering
\caption{Comparison of dynamic definition performance and explainability with state-of-the-art methods on OpenDef-Bench, XD-Violence and HIVAU-70k. T denotes Training and TF denotes Training-free.}
\label{tab:exp_main}
\setlength{\tabcolsep}{3pt}
\resizebox{\columnwidth}{!}{%
\begin{tabular}{lcc|cccc|cc|c}
\toprule
\multirow{3}{*}{Model}
& \multirow{3}{*}{Type}
& \multirow{3}{*}{Online}
& \multicolumn{4}{c|}{OpenDef-Bench}
& \multicolumn{2}{c|}{XD-Violence}
& HIVAU-70k \\
& & & \multicolumn{3}{c}{\shortstack{Temporal\\Localization}} & \multicolumn{1}{c|}{\shortstack{Category\\Selection}} & \multicolumn{2}{c|}{Drift@5} & \shortstack{Video\\Description} \\
\cmidrule(lr){4-6}\cmidrule(lr){7-7}\cmidrule(lr){8-9}\cmidrule(lr){10-10}
& & & AUC & AP & LaAP & ACC & AUC & AP & Win Rate \\
\midrule
HolmesVAU~\cite{Holmes-VAU} & T & $\times$ & 46.35 & 3.74 & -- & 0.71 & -- & -- & -- \\
LaGoVAD~\cite{liu2025lagovad} & T & $\times$ & 48.19 & 6.40 & 6.43 & 9.34 & 85.7 & 37.1 & -- \\
LAVAD~\cite{LAVAD} & TF & $\times$ & 50.13 & 4.91 & 4.84 & -- & 81.7 & 34.8 & -- \\
Qwen3-VL-8B & TF & $\times$ & 54.86 & 10.22 & 7.02 & 12.74 & 78.8 & 34.0 & 61.7\% \\
Qwen3-VL-8B & TF & $\checkmark$ & 52.73 & 8.49 & 6.62 & 27.35 & 81.7 & 24.2 & 52.8\% \\
\textbf{ESOM} & TF & $\checkmark$ & \textbf{59.22} & \textbf{15.88} & \textbf{10.17} & \textbf{32.74} & \textbf{89.3} & \textbf{44.8} & \textbf{69.6\%} \\
\bottomrule
\end{tabular}
}
\end{table}

\begin{table}[t]
\centering
\small
\setlength{\tabcolsep}{4pt}
\caption{Comparison of inference efficiency. FPS denotes the video sampling rate.}
\label{tab:exp_efficiency}
\resizebox{\columnwidth}{!}{%
\begin{tabular}{lccccc cc}
\toprule
\multirow{2}{*}{Model}
& \multirow{2}{*}{\shortstack{Single\\Model}}
& \multirow{2}{*}{Param.}
& \multirow{2}{*}{FPS}
& \multirow{2}{*}{RTF $\downarrow$}
& \multicolumn{1}{c}{UCF-Crime}
& \multicolumn{2}{c}{OpenDef-Bench} \\
\cmidrule(lr){6-6} \cmidrule(lr){7-8}
& & & & & AUC $\uparrow$ & AUC $\uparrow$ & AP $\uparrow$\\
\midrule
LAVAD~\cite{LAVAD} & $\times$ & $\approx$43B & 2 & 11.05 & 79.21 & 50.13 & 4.91 \\
VERA~\cite{vera} & $\times$ & $\approx$9B & 2 & 5.74 & 75.27 & -- & -- \\
ESOM & $\checkmark$ & 8B & 2 & 0.27 & 82.13 & 57.84 & 15.43 \\
ESOM & $\checkmark$ & 8B & 4 & 0.52 & \textbf{86.18} & \textbf{59.22} & \textbf{15.88} \\
\bottomrule
\end{tabular}%
}
\end{table}

In the experiments, we consider both offline and online settings. The offline setting processes the entire video at once, whereas the online setting performs streaming video understanding incrementally.
Based on these two settings, we further construct Qwen3-VL-8B baselines for comparison. The offline baseline uniformly samples 240 frames, whereas the online baseline follows the same sliding-window configuration as ESOM, directly uses category names as prompts, and extracts abnormal intervals from the generated text.

Table~\ref{tab:exp_zeroshot} compares ESOM under the traditional zero-shot setting. To ensure fairness, an online version of VERA is constructed by removing its score post-processing step. 
ESOM achieves the best performance in both temporal localization and fine-grained multi-class classification.

Table~\ref{tab:exp_main} compares performance under dynamic-definition and explainability. HolmesVAU \cite{Holmes-VAU} suffers from strong bias toward seen categories due to its training-data distribution and therefore tends to predict most videos as normal. LaGoVAD \cite{liu2025lagovad} benefits from its dynamic paradigm, but its performance is still limited by the generalization of small models. Other training-free baselines also perform unsatisfactorily. 
Moreover, both offline and online Qwen3-VL-8B baselines achieve inferior performance, indicating that a general-purpose MLLM cannot be directly applied to this task effectively.
By contrast, ESOM consistently achieves the best overall performance.

Table~\ref{tab:exp_efficiency} further compares the efficiency of different models. The results indicate that ESOM achieves superior performance while satisfying real-time requirements. Moreover, ESOM uses only a single model, resulting in a smaller parameter size and making it easier to deploy.

\subsection{Ablation Studies}
We report the main ablations in this section. For more results and visualizations, please refer to the supplementary material.

\subsubsection{Module Effectiveness}
We report ablation studies of modules in Table~\ref{tab:abl_modules}, where the running time is measured with a total token length of 40K. The results show that removing either IIM or HSM leads to a significant decrease in efficiency, accompanied by a noticeable drop in performance,
which indicates that token compression reduces computational overhead while suppressing redundant visual tokens that introduce noise. Meanwhile, the hybrid streaming memory improves performance by compressing weakly related long-term information while maintaining the continuous propagation of short-term information.
In addition, the Probabilistic Scoring module has a substantial impact on temporal localization performance, while the Definition Normalization module plays a critical role in classification performance.

\begin{table}[t]
\centering
\small
\setlength{\tabcolsep}{4pt}
\caption{Ablation study of each module.}
\label{tab:abl_modules}
\resizebox{\columnwidth}{!}{%
\begin{tabular}{lccccccc}
\toprule
\multirow{2}{*}{Method}
& \multicolumn{2}{c}{UCF-Crime}
& \multicolumn{2}{c}{OpenDef-Bench}
& \multirow{2}{*}{\shortstack{Token\\Ratio (\%)}}
& \multirow{2}{*}{\shortstack{Prefill\\Time (s)}}
& \multirow{2}{*}{\shortstack{Decode\\Time (s)}} \\
\cmidrule(lr){2-3} \cmidrule(lr){4-5}
& AUC $\uparrow$ & F1 $\uparrow$ & AUC $\uparrow$ & ACC $\uparrow$ & & & \\
\midrule
ESOM & \textbf{86.18} & \textbf{41.26} & \textbf{59.22} & \textbf{32.74} & \textbf{0.22} & \textbf{0.381} & \textbf{8.31} \\
$-$ w/o IIM 
& 85.75 & 40.71 & 58.70 & 31.65 & 0.44 & 0.865 & 10.94 \\
$\quad -$ w/o HSM
& 80.53 & 40.70 & 54.87 & 31.86 & 1.00 & 2.692 & 18.07 \\
$\qquad -$ w/o PS
& 76.11 & 40.70 & 52.41 & 31.86 & 1.00 & 2.692 & 18.07 \\
$\quad \qquad -$ w/o DN
& 71.99 & 35.63 & 52.73 & 27.35 & 1.00 & 2.692 & 18.07 \\
\bottomrule
\end{tabular}%
}
\end{table}

\subsubsection{Comparison with Other Token Compression Methods}

\begin{figure}[t!]
    \centering
    \includegraphics[width=0.8\linewidth]{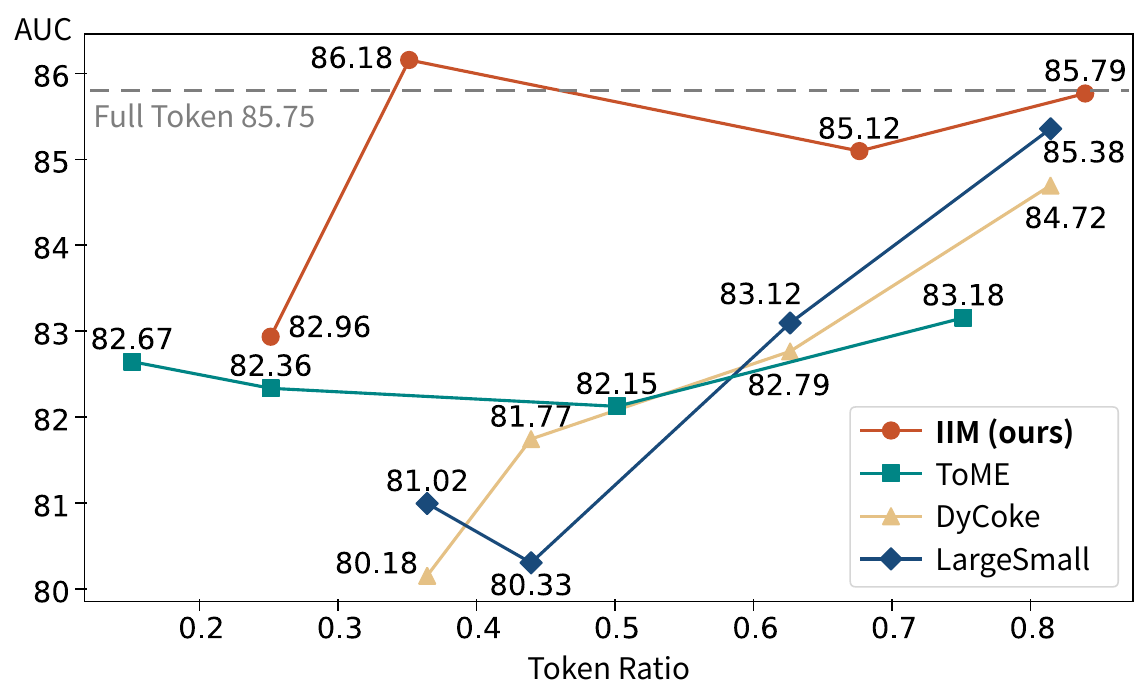}
    \caption{Comparison of different token compression methods \cite{tome,dycoke,keye} under varying token ratios on UCF-Crime.}
    \label{fig:abl_compress_ratio}
\end{figure}

To further validate the effectiveness of the token compression module, we replace it with several general-purpose alternatives and compare their performance under different compression ratios. Specifically, ToME~\cite{tome} is applied independently to each frame, DyCoke refers to the Stage 1 strategy in \cite{dycoke}, and LargeSmall denotes the high–low resolution interleaving strategy used in \cite{keye}. As shown in Fig.~\ref{fig:abl_compress_ratio}, IIM consistently delivers the best performance across all compression ratios.

\subsubsection{Different Backbones}

We also evaluate different MLLMs within our framework in Table~\ref{tab:model_cmp}, including Qwen2.5-VL \cite{qwen2.5-vl}, Qwen3-VL \cite{qwen3-vl}, and Keye-VL-1.5 \cite{keye}. Among them, Qwen3-VL achieves the best temporal performance, and larger models tend to perform better on classification, but do not necessarily show stronger temporal localization ability. Considering efficiency, we finally adopt the 8B variant.

\begin{table}[t]
\centering
\setlength{\tabcolsep}{4pt}

\begin{minipage}[t]{0.4\columnwidth}
\centering
\scriptsize
\caption{Comparison of different backbones}
\label{tab:model_cmp}

\vspace{0.3em}

\resizebox{\linewidth}{!}{%
\begin{tabular}{lcc}
\toprule
\multirow{2}{*}{Method} 
& \multicolumn{2}{c}{UCF-Crime} \\
\cmidrule(lr){2-3}
& AUC $\uparrow$ & F1 $\uparrow$\\
\midrule
Qwen3-VL-4B      & 84.62 & 40.18 \\
Qwen3-VL-8B      & \textbf{86.18} & 41.26 \\
Qwen3-VL-32B     & 84.74 & \textbf{43.18} \\
Qwen2.5-VL-7B    & 73.06 & 38.62 \\
Keye-VL-1.5-8B   & 81.39 & 20.74 \\
\bottomrule
\end{tabular}%
}
\end{minipage}
\hfill
\begin{minipage}[t]{0.57\columnwidth}
\centering
\scriptsize
\caption{Ablation studies on the definition normalization module.}
\label{tab:abl_dn}

\vspace{0.3em}

\renewcommand{\arraystretch}{1.05}
\setlength{\tabcolsep}{3pt}
\resizebox{\linewidth}{!}{%
\begin{tabular}{ccccc}
\toprule
\multirow{2}{*}{\shortstack{Anonymous\\Category}} 
& \multirow{2}{*}{\shortstack{Start--End\\Definition}} 
& \multirow{2}{*}{Format} 
& \multicolumn{2}{c}{UCF-Crime} \\
\cmidrule(lr){4-5}
& & & AUC $\uparrow$ & F1 $\uparrow$ \\
\midrule
$\checkmark$ & $\checkmark$ & markdown & \textbf{86.18} & \textbf{41.26} \\
$\checkmark$ & $\checkmark$ & json     & 84.27 & 41.08 \\
$\checkmark$ & $\checkmark$ & tsv      & 84.17 & 43.01 \\
$\checkmark$ & $\times$     & markdown & 83.59 & 35.23 \\
$\times$     & $\checkmark$ & markdown & 83.26 & 37.48 \\
\bottomrule
\end{tabular}%
}
\end{minipage}

\end{table}

\subsubsection{Effectiveness of the Designs in DN}

Table~\ref{tab:abl_dn} presents ablation studies on the individual components of the DN module, comparing whether anonymous category names are used, whether start–end boundaries are explicitly defined, and the prompt format used for the definitions. The results confirm the critical role of category anonymization and boundary specification, and indicate that the markdown table format achieves the best overall performance.

\subsubsection{Effectiveness of the Designs in IIM}
Table~\ref{tab:abl_iim} explores the GoF strategy and the P-frame interval in the IIM module. The results show that, within a certain range, increasing the window size and enlarging the P-frame interval can improve the compression ratio while maintaining strong detection performance. An overlarge GoF size leads to long matching intervals, making it hard to accurately find redundant tokens. 

\subsection{Visualizations}
\begin{figure}[t!]
    \centering
    \includegraphics[width=1\linewidth]{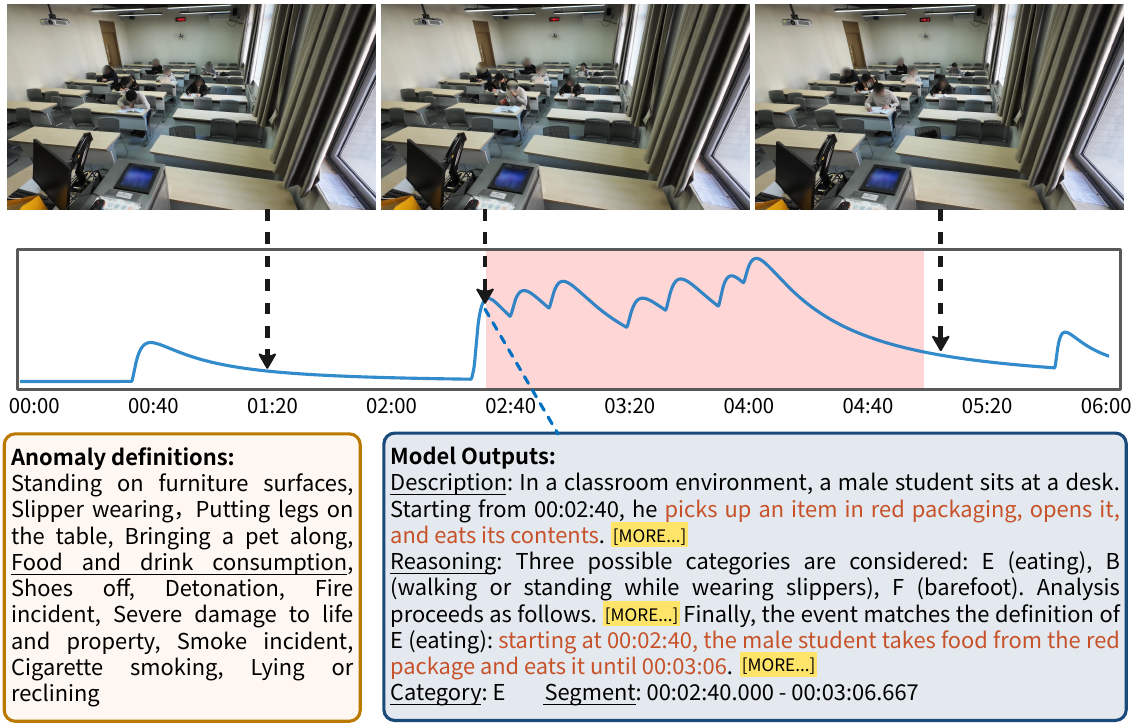}
    \caption{Qualitative results of ESOM. Due to space limitations, some text is truncated and marked with \colorbox[HTML]{FFE46A}{[MORE...]}.}
    \label{fig:result_vis_mini}
\end{figure}

Fig.~\ref{fig:result_vis_mini} presents qualitative examples showing that ESOM can accurately localize abnormal events under given anomaly definitions and produce detailed textual explanations. 
Additional qualitative results are provided in the supplementary material.

\begin{table}[t]
\centering
\caption{Ablation study of the IIM module.}
\label{tab:abl_iim}
\resizebox{0.6\linewidth}{!}{%
\begin{tabular}{cccc}
\toprule
\multirow{2}{*}{GoF Size} & \multirow{2}{*}{P-frame Interval} 
& \multicolumn{2}{c}{UCF-Crime} \\
\cmidrule(lr){3-4}
& & AUC $\uparrow$ & F1 $\uparrow$ \\
\midrule
15 & 7 & 84.31 & 40.93 \\
8 & 7 & \textbf{86.18} & 41.26 \\
4 & 3 & 83.81 & 38.46 \\
4 & 1 & 86.00 & \textbf{41.93} \\
\bottomrule
\end{tabular}
}
\end{table}

\section{Conclusion}
In this paper, we propose ESOM, a novel framework for open-world video anomaly detection.
We first build a training-free framework that can address the challenges of generalization, output format, and dynamic definitions in the open-world setting. 
This framework includes a Definition Normalization module to accurately adapt to dynamic anomaly definitions, and a Probabilistic Scoring module to produce frame-level scores rather than interval-level outputs.
We further propose the IIM and HSM modules to improve efficiency. By reducing spatio-temporal redundancy and minimizing interference with the original inference process, these modules reduce 78\% of tokens while maintaining performance.
In addition, we construct a new benchmark, OpenDef-Bench, which supports evaluations under diverse anomaly definitions.
Experiments demonstrate that ESOM achieves SOTA performance while supporting real-time and online inference.
Future work can explore accurate detection using models with fewer parameters and higher compression ratios.

{
    \bibliographystyle{IEEEtran}
    \bibliography{main}
}

\newpage

\vfill

\end{document}